\documentclass[11pt]{article}

\usepackage[final]{acl}

\usepackage{times}
\usepackage{latexsym}
\usepackage{array}
\usepackage{paralist} 
\usepackage{float}

\usepackage[T1]{fontenc}

\usepackage[utf8]{inputenc}

\usepackage{microtype}

\usepackage{inconsolata}

\usepackage{graphicx}
\usepackage{linguex}
\usepackage{booktabs}
\usepackage{amsmath}
\usepackage{annotates}

\newcommand{\new}[1]{{#1}}

\newcommand{\exponeresults}{
\begin{figure*}[th!]
  \centering
  \includegraphics[width=\textwidth]{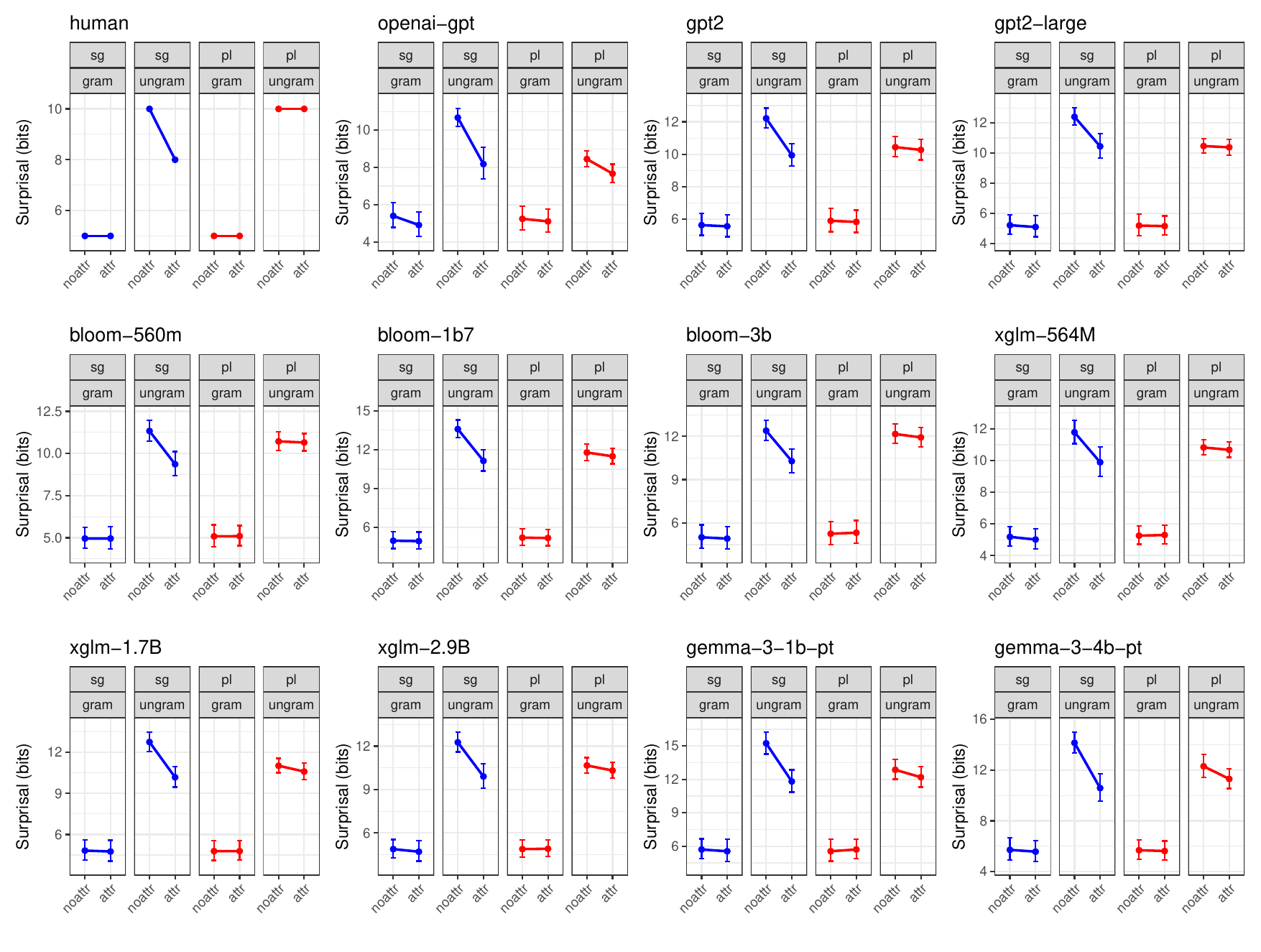}  
  \vspace*{-1.5em}
  \caption{
    Surprisal predictions across models and conditions for Experiment~1. Top left panel: 
    Schematic human performance.
    Each sub-panel shows the effect of attraction in one of the four relevant configurations (Table~\ref{tab:exp1-materials}): 
    Singular subjects (sg) in sub-panels 1–2;  plural (pl) in sub-panels 3–4.
    Geometric means, confidence intervals 95\%.
  }
  \label{fig:exp1-results}
\end{figure*}

\begin{table*}[t!]
  \centering
  \begin{tabular}{llccccccccccccc}
    \toprule
    \multicolumn{2}{l}{}
      & \multicolumn{2}{c}{\textbf{Reference}}
      & \multicolumn{3}{c}{\textbf{GPT}}
      & \multicolumn{3}{c}{\textbf{Bloom}}
      & \multicolumn{3}{c}{\textbf{XGLM}}
      & \multicolumn{2}{c}{\textbf{Gemma-3}} \\
    \cmidrule(lr){3-4} \cmidrule(lr){5-7} \cmidrule(lr){8-10} \cmidrule(lr){11-13} \cmidrule(lr){14-15}
    \textbf{Subject} & \textbf{Gramm.}
      & \rotatebox{90}{Human} & \rotatebox{90}{Theor.\ poss.}
      & \rotatebox{90}{120M} & \rotatebox{90}{2 (127M)} & \rotatebox{90}{2 (744M)}
      & \rotatebox{90}{560M} & \rotatebox{90}{1.7B} & \rotatebox{90}{3B}
      & \rotatebox{90}{564M} & \rotatebox{90}{1.7B} & \rotatebox{90}{2.9B}
      & \rotatebox{90}{1B} & \rotatebox{90}{4B} \\
    \midrule
    singular & gramm.
      &                       & \color{red}$\nearrow$
      & $\searrow$            &                       &
      &                       &                       &
      & $\searrow$            &                       &
      &                       &            \\
    singular & ungramm.
      & \color{red}$\searrow$ & \color{red}$\searrow$
      & $\searrow$            & $\searrow$            & $\searrow$
      & $\searrow$            & $\searrow$            & $\searrow$
      & $\searrow$            & $\searrow$            & $\searrow$
      & $\searrow$            & $\searrow$ \\
    plural & gramm.
      &                       & \color{red}$\nearrow$
      &                       &                       &
      &                       &                       &
      &                       &                       &
      &                       &            \\
    plural & ungramm.
      &                       & \color{red}$\searrow$
      & $\searrow$            & $\searrow$            &
      &                       & $\searrow$            & $\searrow$
      & $\searrow$            & $\searrow$            & $\searrow$
      & $\searrow$            & $\searrow$ \\
    \bottomrule
  \end{tabular}
  \caption{Experiment~1: Observed and predicted agreement attraction effects for prepositional modifiers.
    \textit{Gramm.}: (un)grammatical condition.
    Arrows indicate significant effects: upward = inhibitory, downward = facilitatory.
    \textit{Theor.\ poss.}: effects that are explainable in terms of agreement attraction.}
  \label{tab:exp1-results}
\end{table*}
}

\newcommand{\exptworesults}{
\begin{figure*}[t!]
  \centering
  \includegraphics[width=\textwidth]{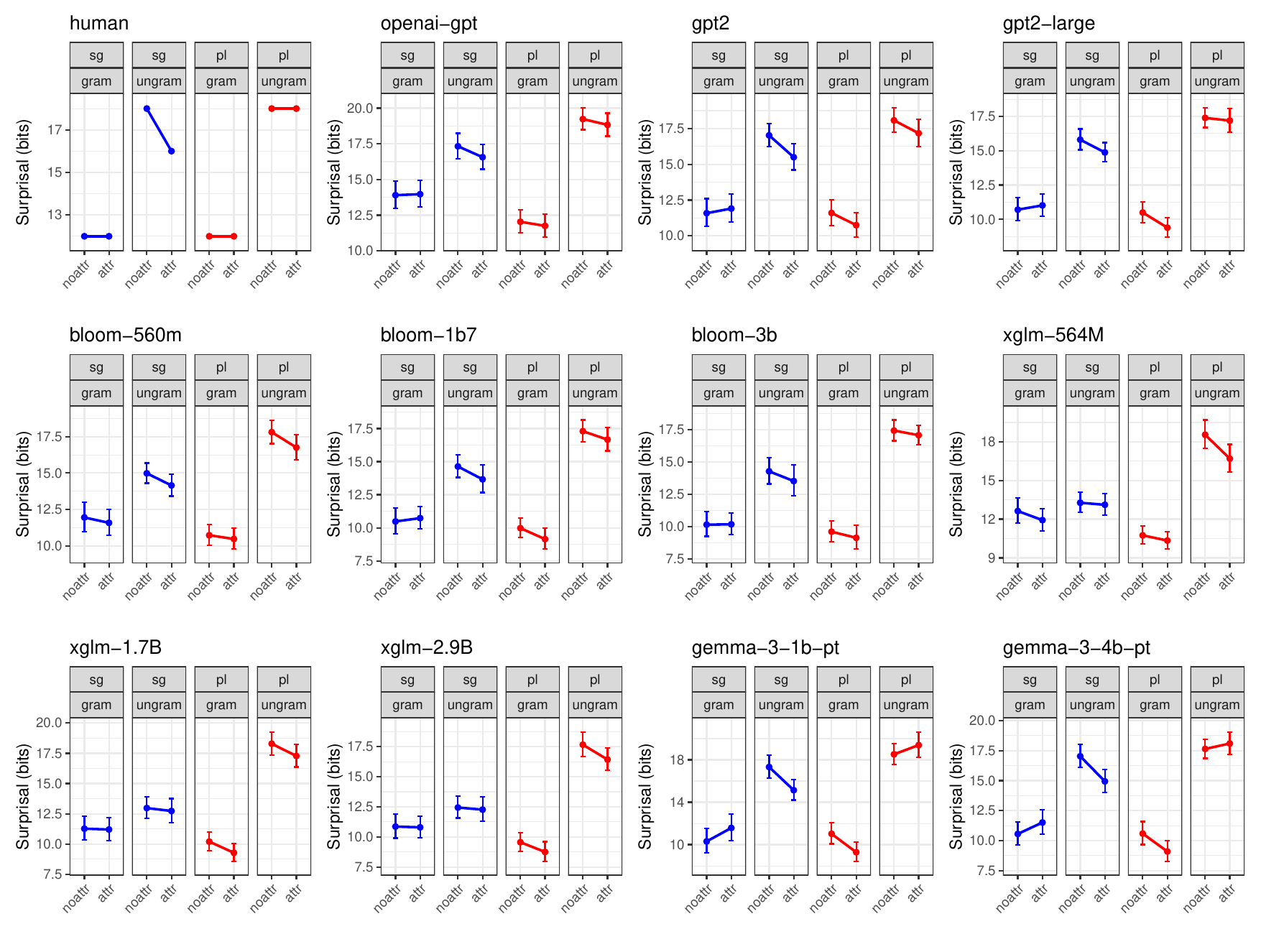}  
  \vspace*{-2em}
  \caption{Surprisal predictions across models and conditions for Experiment~2. Top left panel: Schematic human performance.
       Sub-panels show attraction effects in one of the four relevant configurations (able~\ref{tab:exp2-materials}): Singular RC-subjects (sg) in sub-panels 1–2; plural (pl) in sub-panels 3–4.
    Geometric means, confidence intervals 95\%.
  }
  \label{fig:exp2-results}
\end{figure*}

\begin{table*}[t!]
  \centering
  \begin{tabular}{llccccccccccccc}
    \toprule
    \multicolumn{2}{l}{}
      & \multicolumn{2}{c}{\textbf{Reference}}
      & \multicolumn{3}{c}{\textbf{GPT}}
      & \multicolumn{3}{c}{\textbf{Bloom}}
      & \multicolumn{3}{c}{\textbf{XGLM}}
      & \multicolumn{2}{c}{\textbf{Gemma-3}} \\
    \cmidrule(lr){3-4} \cmidrule(lr){5-7} \cmidrule(lr){8-10} \cmidrule(lr){11-13} \cmidrule(lr){14-15}
    \textbf{RC subject} & \textbf{Gramm.}
      & \rotatebox{90}{Human} & \rotatebox{90}{Theor.\ poss.}
      & \rotatebox{90}{120M} & \rotatebox{90}{2 (127M)} & \rotatebox{90}{2 (744M)}
      & \rotatebox{90}{560M} & \rotatebox{90}{1.7B} & \rotatebox{90}{3B}
      & \rotatebox{90}{564M} & \rotatebox{90}{1.7B} & \rotatebox{90}{2.9B}
      & \rotatebox{90}{1B} & \rotatebox{90}{4B} \\
    \midrule
    singular & gramm.
      &                       & \color{red}$\nearrow$
      &                       &                       &
      & $\searrow$            &                       &
      & $\searrow$            &                       &
      & $\nearrow$            & $\nearrow$ \\
    singular & ungramm.
      & \color{red}$\searrow$ & \color{red}$\searrow$
      & $\searrow$            & $\searrow$            & $\searrow$
      & $\searrow$            & $\searrow$            & $\searrow$
      &                       &                       &
      & $\searrow$            & $\searrow$ \\
    plural & gramm.
      &                       & \color{red}$\nearrow$
      & $\searrow$            & $\searrow$            & $\searrow$
      &                       & $\searrow$            & $\searrow$
      & $\searrow$            & $\searrow$            & $\searrow$
      & $\searrow$            & $\searrow$ \\
    plural & ungramm.
      &                       & \color{red}$\searrow$
      & $\searrow$            & $\searrow$            &
      & $\searrow$            & $\searrow$            &
      & $\searrow$            & $\searrow$            & $\searrow$
      & $\nearrow$            &            \\
    \bottomrule
  \end{tabular}
  \caption{Experiment~2: Observed and predicted agreement attraction effects for relative clauses.
    \textit{Gramm.}: (un)grammatical condition.
    Arrows indicate significant effects: upward = inhibitory, downward = facilitatory.
    \textit{Theor.\ poss.}: effects that are explainable by agreement attraction.
    NB.\ All predicted effects in the plural/grammatical configuration are in the direction that cannot be explained.}
  \label{tab:exp2-results}
\end{table*}
}

\title{Diverging Transformer Predictions for Human Sentence Processing: \\ A Comprehensive Analysis of Agreement Attraction Effects}

\author{Titus von der Malsburg \and Sebastian Padó \\
  University of Stuttgart, Germany \\
 \texttt{\{titus.von-der-malsburg@ling, pado@ims\}.uni-stuttgart.de}
}

\begin{document}
\newpage

\maketitle
\begin{abstract}
  Transformers underlie almost all state-of-the-art language models in computational linguistics, yet their cognitive adequacy as models of human sentence processing remains disputed.
  In this work, we use a surprisal-based linking mechanism to systematically evaluate eleven autoregressive transformers of varying sizes and architectures on a more comprehensive set of English agreement attraction configurations than prior work.
  Our experiments yield mixed results:
  While transformer predictions generally align with human reading time data for prepositional phrase configurations, performance degrades significantly on object-extracted relative clause configurations.
  In the latter case, predictions also diverge markedly across models, and no model successfully replicates the asymmetric interference patterns observed in humans.
  We conclude that current transformer models do not explain human morphosyntactic processing, and that evaluations of transformers as cognitive models must adopt rigorous, comprehensive experimental designs to avoid spurious generalizations from isolated syntactic configurations or individual models. 
\end{abstract}

\section{Introduction}


Neural network architectures now dominate both applied and theoretical approaches to modeling linguistic phenomena \citep{LinzenBaroni2021}.
Specifically, \textit{transformers} \citep{vaswani2017attention} have emerged as a major success story, supplanting an earlier generation of recurrent network architectures by offering improved scalability and parallelization, and, on many tasks, greater parameter efficiency (but see \citealt{BeckEtAl2024}).
Virtually all state-of-the-art large language models (LLMs) are now based on transformers \citep{NEURIPS2020_1457c0d6}.

From a psycholinguistic point of view, the success of transformers stands in contrast to the simplistic next-word prediction tasks on which
they are trained.
This prompts the question: To what extent do transformer models capture the cognitive processes underlying human language processing? Historically, neural network models have been successfully leveraged to account for a wide range of findings in human language comprehension and production, including patterns in reaction times \citep{Wilcox2020OnTPA}, event-related potentials \citep{frank-2024-neural}, and functional MRI activity \citep{Toneva2019InterpretingAIA} during sentence comprehension.
These results raised expectations about the potential of transformers as cognitive models of human language processing \citep{CuskleyEtAl2024}.

In this paper, we focus on \textit{reading time studies} which constitute a central methodology in psycholinguistics due to their sensitivity and richness as measures of on-line processing effort.
A number of studies have tested the ability of transformer-based LLMs to account for human reading times on benchmark phenomena such as garden path sentences \citep{amouyal-etal-2025-lm}.
A phenomenon that leads to a particularly intriguing pattern of reading times is \textit{agreement attraction} \citep{BockMiller1991}.
Agreement attraction is the cross-linguistic phenomenon where the processing of a verb is impeded by a grammatically unrelated noun.
Consider the following examples:

\ex.
\a. *The key were rusty.
\b. *The key to the cabinet were rusty.
\c. *The key to the cabinets were rusty.

From the standpoint of linguistic theory, all three sentences in \Last are equally ungrammatical due to a mismatch of the verb “were” (plural) and  its subject “key” (singular).
Nonetheless, experiments have shown that \Last[c] is perceived by readers to be more acceptable, and indeed the verb region is read faster than in \Last[b], suggesting reduced processing difficulty \citep{WagersEtAl2009}.\footnote{In generation, a similar effect takes place, in that speakers utter \Last[c] more frequently and easily than \Last[b].}
This effect can be attributed to the intervening noun “cabinets”, which is available to (incorrectly) satisfy the verb’s demand for a plural subject \cite[e.g.,][and many others]{PEARLMUTTER1999427, WagersEtAl2009, LagoEtAl2015, LaurinavichyuteMalsburg2022}.

Agreement attraction effects are currently best explained by the specific ways in which linguistic information is stored and accessed in human working memory.
Consequently, agreement attraction presents an interesting benchmark for LLMs, as it is unclear to what extent transformers would exhibit similar behavior at this level of specificity.

A number of previous studies have evaluated transformers on agreement attraction datasets \citep[e.g.,][]{BazhukovEtAl2024CoNLL, RyuLewis2021CMCL, TimkeyLinzen2023EMNLP} and have reported mostly positive results, finding that transformer predictions aligned well with human reading times (cf.\ Section \ref{sec:relatedwork} for details).
While encouraging, these studies were limited in multiple ways:

\begin{enumerate}

\item
  Prior studies tested only the construction where the distractor noun is embedded in a prepositional modifier phrase (cf.\ Table~\ref{tab:exp1-materials}).
  However, agreement attraction effects have also been observed in a range of other constructions, most importantly in  object-extracted relative clauses (cf.\ Table~\ref{tab:exp2-materials}).
  Transformer predictions for that construction are, to our knowledge, untested, and it is not safe to assume that effects are necessarily the same as for prepositional modifiers.

\item
  Prior studies only tested configurations in which the agreement attraction effect is observed.
  However, human studies such as \citet{WagersEtAl2009} have established that attraction effects appear reliably only if the verb’s subject is singular and the sentence ungrammatical.
  That is, attraction effects are believed to be largely absent
  when the subject is plural or when the sentence is grammatical.
\new{While we cannot rule out that some null effects reported in the literature are in fact not truly null \citep{LaurinavichyuteMalsburg2024}, what is crucial is the direction and the relative magnitudes of effects, on which  there is overwhelming agreement in the literature. 
Arguably, it is this whole pattern of effects that any account of agreement attraction -- transformer-based or not -- has to explain in order to qualify as a scientifically satisfying explanation of agreement attraction.}
  However, previous studies have not confirmed the absence of agreement attraction effects where they should not appear.




\item
  Prior studies considered individual transformer models (typically GPT-2), leaving open whether their findings are specific to that model or apply to transformers more broadly -- a crucial gap in generalization.

\item
  Finally, some of the existing studies considered models from the BERT family \cite{devlin-etal-2019-bert} which are a priori cognitively implausible due to their bi-directional nature.

\end{enumerate}

\noindent Our study addresses these limitations by testing a range of autoregressive transformer models, carrying out a full sweep of all theoretically relevant syntactic configurations.
These configurations manipulate the number of the subject, grammaticality, and potential attraction through the number feature of a distractor noun.
To test the generalizability of findings beyond prepositional modifier constructions (Table~\ref{tab:exp1-materials}), we also test object-extracted relative clause constructions (Table~\ref{tab:exp2-materials}).
Finally, we consider a range of 11 transformer models from four different open-weight families, both monolingual and multilingual.
To systematically disentangle the contributions of the various factors, we analyze the LLM predictions using regression modeling.

Our findings show that transformer predictions are consistent and align well with human data for the English prepositional modifier construction, consistent with earlier research.
Crucially, however, we also find that transformer predictions diverge in the case of English object-extracted relative clauses, a result that casts serious doubt on transformers as viable models of human sentence processing.
On a methodological level, we conclude that cognitive evaluations of computational models must adopt comprehensive experimental designs
to avoid spurious generalizations from isolated syntactic configurations or individual models.

\section{Related Work}
\label{sec:relatedwork}

Agreement attraction has been identified early on as an important test case for neural language models because it reflects the peculiar way in which human incremental sentence processing proceeds as opposed to what would be expected purely based on linguistic theory.
Early work by \citet{LinzenEtAl2016} established that Long Short-Term Memory networks can successfully predict verb number across intervening distractors if specifically trained to do so \citep[see also][]{LinzenLeonard2018CogSci}.
Crucially, though, they also showed that LSTM performance was consistently degraded in the presence of plural attractors mirroring human attraction effects.
This result was later generalized to other neural models \citep{BernardyLappin2017}, showing that neural LMs more generally naturally exhibit human-like processing patterns.
Subsequently, \citet{GulordavaEtAl2018} replicated these findings with LSTMs that were trained only on next-word prediction \citep[cf.][]{ArehalliLinzen2020CogSci}.

Against this backdrop, it is unsurprising that early successes of neural language models, particularly LSTMs, at capturing human agreement attraction effects generated significant interest in such models as candidates for explaining human language processing \citep{LinzenBaroni2021}.

These early findings effectively started a new research area that investigates these models’ psycholinguistic adequacy, and
when transformer-based LMs appeared, they fueled
expectations that these models might account even better for human sentence processing.
Indeed, several studies appeared to confirm this promise
particularly in capturing reading time patterns such as those seen in agreement attraction \citep{RyuLewis2021CMCL, BazhukovEtAl2024CoNLL, ZacharopoulosEtAl2023EMNLP, KimDavis2025SCiL}.
However, recent work again casts doubt on LMs’ ability to fully account for human syntactic processing.
Specifically, studies on syntactic disambiguation demonstrate that single-stage language models and transformer-based surprisal specifically under-predict the magnitude of difficulty for garden-path sentences \citep{VanSchijndelLinzen2021, HuangEtAl2024}.
Our study extends this critical perspective to the apparently well-established agreement attraction results.

\section{Method}

\paragraph{Study Design.}

We report experiments on two different syntactic realizations of agreement attraction that have previously been studied in the psycholinguistic agreement attraction literature, namely prepositional modifier constructions (Experiment~1) and object-extracted relative constructions (Experiment~2).
Both experiments tested eight configurations (2×2×2): with and without distracting nouns, with singular and plural subjects, and with verbs that match or mismatch the number of the subject.
As human reference, we used self-paced reading times from experiments 3 and 4 reported by \citet{WagersEtAl2009}.
This study provides, to date, the most comprehensive benchmark for agreement attraction in English: Experiment~3 tested 384 sentences with 60 participants, Experiment~4 tested 192 sentences with 46 participants. The study reports aggregated reading times across all participants. 

\paragraph{Simulation.} We compared human reading times against reading time predictions produced on the basis of LLMs.
Specifically, we passed all stimuli through autoregressive language models (see below for model selection criteria) which processed the stimuli in a strict left-to-right fashion.
This is a plausible setup to model the outcomes of a self-paced reading study, where participants are also limited to strict left-to-right reading.

We followed the established surprisal theory of reading time prediction  \citep{LEVY20081126, DembergKeller2008, BostonEtAl2008, fernandez-monsalve-etal-2012-lexical} which measures the 'surprisal' of some token $w_i$ as its negative log probability given the preceding context $w_0\dots w_{i-1}$:
\[ \text{surprisal [bits]} = - \log_2 P(w_i | w_0 \dots w_{i-1})  \]
Log probability falls directly out of any autoregressive language model.
The surprisal is then assumed to be linearly related to reading time \citep{SmithLevy2013, ShainEtAl2024}.
In both experiments, we compared human and LLM-based reading times at a specific word of interest, namely the verb that could potentially be affected by attraction, marked in bold in Table~\ref{tab:exp1-materials} (Experiment~1) and Table~\ref{tab:exp2-materials} (Experiment~2), respectively.

\paragraph{LLM choice.}
\new{We are primarily interested in the extent to which general language models, trained with the standard
next-word prediction task, are able to account for human agreement attraction patterns. For this reason, we only consider foundation models without instruction tuning, and do not carry out task-specific tuning of models \citep{LinzenBaroni2021,yoshida2026existenceproofneurallanguage}.}
Furthermore, we restricted the analysis to current 'mid-sized' autoregressive transformer models with parameter counts between 124M and 4B, following prior work that found better matches with human reading times for such models compared to both smaller and larger models \citep{OhSchuler2023}.
We wanted to assess the effects of both model architecture (or training regimen, respectively, which are all but impossible to disentangle) and model size.
Therefore, we used eleven pre-trained transformers of different sizes from four different families, including monolingual (GPT, GPT-2) and multilingual (Bloom, XGLM, Gemma-3) ones.\footnote{
\new{Our study excludes BERT since we focus on
rapid incremental (i.e., word-by-word) effects in a self-paced reading setup where human participants do not have access to information from the right-hand context. It is not clear how BERT’s bidirectional architecture could be a psychologically plausible modeling choice for this setup. We also exclude LSTMs, which were historically tested for agreement attraction studies, since it is unclear whether a comparison between recent large transformers and LSTMs would be altogether meaningful.}}


Since all models carry out their own sub-word tokenization, the target verb was occasionally split into multiple tokens, especially with XGLM models.
In these cases, we summed the surprisal bits of the constituent tokens to obtain the verb’s total surprisal.
Recent work has demonstrated tokenization bias, such that there may be a surprisal penalty for words split into multiple tokens \citep{LesciEtAl2025ACL}.
While this could potentially confound some of our comparisons, the words used in our stimuli are highly frequent and are therefore rarely split.\footnote{We obtained the same qualitative results when removing all split words (not reported due to space reasons).}

\paragraph{Analysis.}


We analyzed human reading times and model surprisal at the level of conditions, aggregating per-sentence measurements by averaging.
We ran three types of analyses, from fine-grained to global-level.
In all analyses, the dependent variable was surprisal (in bits).

\begin{enumerate}

\item
  To test for the presence or absence of attraction effects in individual models, we used paired t-tests, one for each theoretically relevant contrast between a particular configuration with vs.\ without attraction.

\item
  We ran linear regression models to test for the presence of the two important asymmetries (observed in humans) at the population level (i.e.\ across transformer models):

  \begin{enumerate}

  \item \textit{Singular-plural asymmetry}: agreement attraction appears when the subject is singular but not when it is plural.

  \item \textit{Grammaticality asymmetry}: attraction appears in ungrammatical but not in grammatical sentences.

  \end{enumerate}

  These regression models were fit for all four theoretically relevant configurations and for both syntactic constructions (Tables \ref{tab:exp1-materials} and \ref{tab:exp2-materials}), resulting in eight models overall.

\item
  To capture global tendencies, particularly the effect of models’ parameter count, we also ran two regression models, one for each experiment, covering all eight configurations across all models \cite{Multilevel12}.
  As fixed effects, we included the conditions (grammaticality, subject number, attraction; all coded as treatment contrasts), the number of model parameters ($\log_{10}$, centered), and all interactions of these variables.

\end{enumerate}
In all regression models, we included stimuli and transformer models as crossed random effects, treating models like participants in a human study.
Due to the small number of models and families, we had to simplify random effect structures to ensure reliable convergence.
In the global-level analyses, we tried nesting individual models within their families, but these models did not reliably converge presumably due to the small number of models in each family.
In the per-configuration analyses, models with a maximal random-effects structure did not reliably converge, so we retained by-item and by-model random intercepts only.
We only report the most relevant results in the text and give the full regression results in the Appendices A through C.

\section{Experiment~1: Prepositional Modifier Construction}

\begin{table*}[t!]
    \centering
    \begin{tabular}{llll}
    \toprule
    & Subject \\
          & Number & Grammatical?  & Example \\
    \midrule
    (a/b) & singular       & grammatical   &  The slogan  on the poster(s) unsurprisingly \textbf{was}  designed \dots \\
    (c/d) & singular       & ungrammatical & *The slogan  on the poster(s) unsurprisingly \textbf{were} designed \dots \\
    (e/f) & plural         & grammatical   &  The slogans on the poster(s) unsurprisingly \textbf{was}  designed \dots \\
    (g/h) & plural         & ungrammatical & *The slogans on the poster(s) unsurprisingly \textbf{were} designed \dots \\
\bottomrule
    \end{tabular}
    \caption{The eight conditions of Experiment~1: agreement attraction in prepositional modifier constructions.
      Attraction is theoretically possible when the number of the distractor (here “poster(s)”) differs from subject number.
      In humans, attraction is observed only in (c/d).}
    \label{tab:exp1-materials}
\end{table*}

\paragraph{Design.}

We used 192 sentences (24 sets in 8 conditions) from Experiment~4 in \citet{WagersEtAl2009}.
The structure is shown in Table \ref{tab:exp1-materials}.
 The subject-verb combination, \textit{slogan(s) \dots\ was/were}, appeared in the main clause, and the attractor noun, \textit{poster(s)}, in a prepositional modifier (\textit{on the poster(s)}) attached to the subject.
As described above, factors were number of subject, grammaticality, and attraction (2×2×2 design).
Attraction is theoretically possible when the number of the distractor (here “poster(s)”) differs from subject number.
In humans, attraction is observed only in (c/d).

\paragraph{Results.}

Average surprisal for verbs was 5.5\,bits when they were grammatical and 11.3\,bits when ungrammatical.
Verbs’ surprisal scores were highly correlated across models, with correlations ranging from 0.77 (GPT vs.\ Gemma-3 4B) to 0.98 (between the two larger XGLM models).
Correlations were higher within model families (0.93–0.98).
This confirms that models capture the notion of grammaticality \cite{LinzenBaroni2021}. 

Figure~\ref{fig:exp1-results} shows results for all models.
Each sub-panel represents one of the four theoretically relevant contrasts for agreement attraction (in Table~\ref{tab:exp1-materials}: a/b, c/d, e/f, g/h).
The overall pattern of results was remarkably stable and closely mirrored human performance: pronounced attraction effects were observed only in the ungrammatical configuration with singular subject (2.4\,bits reduction).

Table~\ref{tab:exp1-results} shows the results of the statistical significance tests, comparing them both against theoretically possible and actually observed effects in humans.
Consistent with Figure~\ref{fig:exp1-results}, all models correctly predicted the agreement effect (c/d).
However, all models except GPT2 (744M) and Bloom (560M) also predicted a theoretically plausible but unobserved facilitation effect for ungrammatical plural subjects (4th row), and two models (GPT and XGLM 564M) predicted theoretically impossible minor facilitatory effects in the grammatical singular condition (first row).

\exponeresults



Regarding the singular-plural asymmetry, there was no evidence in grammatical sentences for any population-level agreement attraction effects.
In ungrammatical sentences, there was robust evidence for a facilitatory agreement attraction effect with singular subjects ($\hat\beta$=$-2.4$\,bits, $t$=$-22.9$), and this effect was much reduced when the subject was plural ($\hat\beta$=$2.0$\,bits, $t$=$13.7$).
As for the grammaticality asymmetry, in sentences with singular subjects, surprisal was substantially increased for ungrammatical verbs ($\hat\beta$=$7.2$\,bits, $t$=$67.1$), but this effect was substantially reduced in the presence of attraction ($\hat\beta$=$-2.3$\,bits, $t$=$-14.9$).
In sentences with plural subjects, ungrammatical verbs again had increased surprisal ($\hat\beta$=$5.6$\,bits, $t$=$15.1$), which was only slightly reduced in the presence of attraction ($\hat\beta$=$-0.35$\,bits, $t$=$-2.8$).

In addition, the global-level regression showed that larger models produced slightly higher surprisal for ungrammatical sentences ($\hat\beta$=$1.4$\,bits, $t$=$5.2$).
Other than that, there were no effects of model size.


Overall, we find that transformer-based autoregressive LLMs reliably predict human agreement attraction effects in prepositional modifier constructions, consistent with earlier research.
In addition, we establish that these models (almost always) abstain from predicting effects inconsistent with human data.
The exception is that most models predict facilitatory effects in ungrammatical sentences with plural subjects.
To our knowledge, such effects have never been attested in humans.
However, this configuration has also rarely been tested and there are two studies that found such effects in sentence production which tends to mirror comprehension \citep{FranckEtAl2002, Staub2009}.
It would therefore be premature to rule out such effects and it may be worth revisiting this in future work.

\section{Experiment~2: Object-Extracted Relative Clause Construction}

\begin{table*}[th!]
  \centering
  \begin{tabular}{llll}
    \toprule
    & RC subject number & Grammatical?  & Example \\
    \midrule
    (a/b) & singular        & grammatical   &  The marine(s) who the officer  \textbf{wants} to promote \dots \\
    (c/d) & singular        & ungrammatical & *The marine(s) who the officer  \textbf{want}  to promote \dots \\
    (e/f) & plural          & grammatical   &  The marine(s) who the officers \textbf{want}  to promote \dots \\
    (g/h) & plural          & ungrammatical & *The marine(s) who the officers \textbf{wants} to promote \dots \\
    \bottomrule
  \end{tabular}
  \caption{The eight conditions of Experiment~2: agreement attraction in object-extracted relative clause constructions.
    Attraction is theoretically possible when the number of the distractor (here “marine(s)”) differs from subject number.
    In humans, attraction is observed only in (c/d).}
  \label{tab:exp2-materials}
\end{table*}

\paragraph{Design.}

We used 384 sentences (48 sets) from Experiment~3 in \citet{WagersEtAl2009}.
As Table~\ref{tab:exp2-materials} shows, the structure is parallel to our Experiment~1, with the difference that the subject manipulated in Experiment~2 is the subject of the relative clause (“officer(s)”), and the (previously manipulated) main clause subject (“marine(s)”) was the distractor.

In \citet{WagersEtAl2009}, human readers were remarkably consistent across syntactic phenomena: 
As in the prepositional phrase experiment, attraction was observed only in the contrast (c/d) although other effects are theoretically possible.


\paragraph{Results.}

The average surprisal of grammatical verbs was 11.2\,bits vs.\ 16.5\,bits for ungrammatical verbs.
These values were higher than in Experiment~1 because the verbs in Experiment~2, which are content verbs, are more difficult to predict than the verbs in Experiment~1, which are auxiliaries.
As before, surprisal at the verbs was highly correlated across models, with correlations ranging from 0.68 (GPT vs.\ Bloom 3B) to 0.95 (between the two larger Bloom models) and higher within model families (0.83–0.95).

Figure~\ref{fig:exp2-results} shows surprisal predictions for all models across the eight conditions, and Table~\ref{tab:exp2-results} the corresponding significance tests.
In contrast to Experiment~1, model predictions diverged substantially from each other and from human performance.
While most models (except XGLM) showed at least the signature human-observed effect of agreement attraction (c/d), predictions for the other contrasts were mixed, and models predicted many spurious effects. 
Notably, all models but Bloom (560M) predicted a facilitatory attraction effect for grammatical sentences with plural subject (e/f, 3rd row in Table~\ref{tab:exp2-results}), where theory would predict an inhibitory effect.
For other contrasts, such as (a/b) (first row), models predicted the full range of possible outcomes.
Crucially, none of the models faithfully reproduced the pattern of effects observed in humans or expected based on theory.

\exptworesults
%

As for the singular-plural asymmetry, 
 there was an unexpected facilitatory attraction effect for plural subjects
 in grammatical sentences ($\hat\beta$=$-0.93$\,bits, $t$=$-5.8$).
In ungrammatical sentences, there was an unexpected facilitatory effect regardless of subject number ($\hat\beta$=$-0.9$\,bits, $t$=$-6.6$).

Regarding the grammaticality asymmetry, we found the following:
In sentences with singular subjects, the effect of ungrammaticality was reduced by attraction ($\hat\beta$=$-1.0$\,bits, $t$=$-5.3$, classic agreement attraction effect).
When the subject was plural, there was an unexpected facilitatory effect regardless of grammaticality ($\hat\beta$=$-0.8$\,bits, $t$=$-7.1$).
The global-level regression confirmed these findings and also showed that surprisal was reduced generally in larger models.
There were no other effects of model size.

These results show that transformer-based autoregressive LLMs fall short of offering an explanation for human agreement attraction effects in object-extracted relative clauses.
They neither predict human reading times at the level of individual models, nor does the ensemble of models yield general trends consistent with experimental results or theory.
We also find large differences across models, showing that transformers do not behave uniformly as a class on this particular construction.


\section{Discussion and Conclusions}

Our study investigated the cognitive adequacy of transformer-based autoregressive LLMs by testing their reading time predictions across two well-studied agreement attraction configurations.
Unlike prior studies with LSTMs, we found major differences between the syntactic constructions:
While model predictions were largely consistent and well-aligned with human performance in prepositional modifier constructions (Experiment~1), the results for object-extracted relative clauses are best described as chaotic (Experiment~2):
Models diverged markedly from human performance, and even predicted effects in the opposite direction of what could be theoretically explained.
Even worse, model predictions were inconsistent across model families, with only some consistency within families.
Consistency of predictions is a zeroth-order requirement for any scientific model: if predictions are inconsistent, testing them does not even make sense.
Our findings thus cast significant doubt on the ability of transformers, as a class, to comprehensively account for human morphosyntactic processing.

\new{One possible concern regarding our study is the size of the datasets \citep{WagersEtAl2009} that we simulate. We 
are confident that they are large enough for our purposes: (a), if they were too small, we would expect only 5\% of significant effects. Instead, we found 68\% (ORC) and 50\% (PP-mod), respectively. (b), we would expect false positive effects to be largely random. Instead, we found effects to be highly consistent (Experiment~1) and largely consistent within architectural families (Experiment~2). (c), Human performance is highly variable due to attentional effects, fatigue, motivation, and task adaptation \cite{BAAYEN2017206}, while these factors do not play a role for LMs. As such, LM experiments should have a higher signal-to-noise ratio, and human experimental datasets
should be sufficient in size for an LM-based study.}

One possible explanation for our findings is that the auxiliary verbs in Experiment~1 were easier to predict than the content verbs in Experiment~2, as reflected in lower surprisal values.
Models may therefore have operated with unreliable estimates of the relevant next-word probabilities.
If true, larger models might align better with humans.
Unfortunately, our results suggest the opposite, with the Gemma-3 models (1B, 4B) deviating most dramatically from humans and GPT models approximating them more closely \citep{OhSchuler2023}.
This contrast suggests another explanation, namely that monolingual transformers (which tend to be smaller) model human performance more closely than multilingual transformers.
However, we did not see any evidence for this in Experiment~1 on prepositional modifier constructions.
Nonetheless, this hypothesis is worth testing in future work.

A more speculative explanation appeals to Marr's levels of analysis \citep{Marr1982}.
Humans and transformers can be taken to share a computational-level goal – the incremental prediction of upcoming input – which is precisely what licenses surprisal as a common linking function between them \citep{Hale2001, LEVY20081126}.
They differ, however, at the algorithmic and implementational levels: human agreement attraction is standardly attributed to cue-based retrieval from a capacity-limited memory (cf.\ our Section~1 and \citealt{10.3389/fpsyg.2018.00002}), whereas transformers implement self-attention over the entire context and are not subject to the same memory constraints.
Crucially, implementational (`hardware') differences alone do not entail divergent behavior – this is the working assumption of the entire comparative enterprise – but they make it an open, empirical question which algorithmic-level differences carry consequences at the computational level.
On this view, the contrast between (simpler) prepositional phrase configurations and (more complex) object relative clause configurations marks a candidate boundary between human-like and non-human-like processing, and isolating the mechanisms responsible is a goal for future work.

Finally, we must also consider the possibility that transformers show agreement attraction effects only because such errors appear in training corpora \citep{BockEtAl2006, Pfau2003}.
Transformers would then predict human behavioral patterns not because of human-like inductive biases, they may simply have learned to mimic human surface-level error patterns.
If true, this could explain why LM predictions were more consistent within families. \new{We leave training data analysis as a promising venue for future study.}

At the methodological level, our findings demonstrate that it is unsafe to generalize from individual models (and individual configurations) to transformer-based LMs as a class (and general capabilities).
As a best practice for future studies, we propose: (a) to cover a comprehensive set of syntactic realizations of the phenomenon and adjacent control constructions; (b) to cover a range of plausible language models that differ in architecture, training and size; and (c) to carry out principled statistical analyses at multiple levels as done in this work.

\section*{Limitations}

The present study is restricted to English stimuli and human self-paced reading time data.
Although agreement attraction is a cross-linguistically robust phenomenon (see, e.g., \citealt{LagoEtAl2015}), evaluating the cognitive plausibility of language models requires testing across a typologically diverse set of languages.
Future work must investigate whether the observed performance differences between monolingual and multilingual models persist, particularly when modeling sentence processing in non-English languages for which experimental reading time data is available.
Furthermore, the selection of evaluated models, while spanning multiple families and sizes, is far from exhaustive.
Testing a broader range of models, including strictly monolingual models trained on massive corpora or alternative architectures, is necessary to determine if the failure to capture human-like processing patterns in relative clauses is an inherent limitation of the next-word prediction objective or an artifact of specific architectural choices.

\bibliography{literature}

\newpage

\appendix

\section{Regression Analyses Testing Singular-Plural Asymmetry}
\label{sec:appendix-sgpl}

\subsection{PP-Modifier, Grammatical}

\begin{table}[H]
\centering
\small
\begin{tabular}{lrrr}
\toprule
\textbf{Fixed effects} & \textbf{Estimate} & \textbf{Std.\ Error} & \textbf{$t$ value} \\
\midrule
(Intercept)            & 5.51182           & 0.35612              & 15.478 \\
pl                     & 0.04180           & 0.07201              & 0.580 \\
attr                   & -0.11356          & 0.07201              & -1.577 \\
pl $\times$ attr       & 0.08413           & 0.10184              & 0.826 \\
\bottomrule
\end{tabular}
\caption{Regression results for prepositional modifier construction, grammatical sentences.  Observations: 1056.  Random effects: by-item and by-model random intercepts (item: 24 levels; model: 11 levels); random slopes were omitted because models with a maximal random-effects structure did not reliably converge.}
\label{tab:app-sgpl-pp-gram}
\end{table}

\subsection{PP-Modifier, Ungrammatical}

\begin{table}[H]
\centering
\small
\begin{tabular}{lrrr}
\toprule
\textbf{Fixed effects} & \textbf{Estimate} & \textbf{Std.\ Error} & \textbf{$t$ value} \\
\midrule
(Intercept)            & 12.7320           & 0.4395               & 28.97 \\
pl                     & -1.5696           & 0.1036               & -15.14 \\
attr                   & -2.3755           & 0.1036               & -22.92 \\
pl $\times$ attr       & 2.0008            & 0.1466               & 13.65 \\
\bottomrule
\end{tabular}
\caption{Regression results for prepositional modifier construction, ungrammatical sentences.  Observations: 1056.  Random effects: by-item and by-model random intercepts (item: 24 levels; model: 11 levels); random slopes were omitted because models with a maximal random-effects structure did not reliably converge.}
\label{tab:app-sgpl-pp-ungram}
\end{table}

\subsection{ORC, Grammatical}

\begin{table}[H]
\centering
\small
\begin{tabular}{lrrr}
\toprule
\textbf{Fixed effects} & \textbf{Estimate} & \textbf{Std.\ Error} & \textbf{$t$ value} \\
\midrule
(Intercept)            & 11.8113           & 0.4490               & 26.306 \\
pl                     & -0.8533           & 0.1128               & -7.563 \\
attr                   & 0.1168            & 0.1128               & 1.035 \\
pl $\times$ attr       & -0.9307           & 0.1596               & -5.833 \\
\bottomrule
\end{tabular}
\caption{Regression results for object-extracted relative clauses, grammatical sentences.  Observations: 2112.  Random effects: by-item and by-model random intercepts (item: 48 levels; model: 11 levels); random slopes were omitted because models with a maximal random-effects structure did not reliably converge.}
\label{tab:app-sgpl-orc-gram}
\end{table}

\subsection{ORC, Ungrammatical}

\begin{table}[H]
\centering
\small
\begin{tabular}{lrrr}
\toprule
\textbf{Fixed effects} & \textbf{Estimate} & \textbf{Std.\ Error} & \textbf{$t$ value} \\
\midrule
(Intercept)            & 15.4838           & 0.4439               & 34.880 \\
pl                     & 2.7572            & 0.1388               & 19.859 \\
attr                   & -0.9144           & 0.1388               & -6.586 \\
pl $\times$ attr       & 0.3678            & 0.1963               & 1.873 \\
\bottomrule
\end{tabular}
\caption{Regression results for object-extracted relative clauses, ungrammatical sentences.  Observations: 2112.  Random effects: by-item and by-model random intercepts (item: 48 levels; model: 11 levels); random slopes were omitted because models with a maximal random-effects structure did not reliably converge.}
\label{tab:app-sgpl-orc-ungram}
\end{table}

\section{Regression Analyses Testing Grammaticality Asymmetry}
\label{sec:appendix-gramungram}

\subsection{PP-Modifier, Singular Subject}

\begin{table}[H]
\centering
\small
\begin{tabular}{lrrr}
\toprule
\textbf{Fixed effects} & \textbf{Estimate} & \textbf{Std.\ Error} & \textbf{$t$ value} \\
\midrule
(Intercept)            & 5.5118            & 0.3783               & 14.570 \\
ungram                 & 7.2202            & 0.1076               & 67.082 \\
attr                   & -0.1136           & 0.1076               & -1.055 \\
ungram $\times$ attr   & -2.2620           & 0.1522               & -14.860 \\
\bottomrule
\end{tabular}
\caption{Regression results for prepositional modifier construction, singular subject.  Observations: 1056.  Random effects: by-item and by-model random intercepts (item: 24 levels; model: 11 levels); random slopes were omitted because models with a maximal random-effects structure did not reliably converge.}
\label{tab:app-gram-pp-sg}
\end{table}

\subsection{PP-Modifier, Plural Subject}

\begin{table}[H]
\centering
\small
\begin{tabular}{lrrr}
\toprule
\textbf{Fixed effects} & \textbf{Estimate} & \textbf{Std.\ Error} & \textbf{$t$ value} \\
\midrule
(Intercept)            & 5.55362           & 0.36811              & 15.087 \\
ungram                 & 5.60875           & 0.08834              & 63.491 \\
attr                   & -0.02943          & 0.08834              & -0.333 \\
ungram $\times$ attr   & -0.34526          & 0.12493              & -2.764 \\
\bottomrule
\end{tabular}
\caption{Regression results for prepositional modifier construction, plural subject.  Observations: 1056.  Random effects: by-item and by-model random intercepts (item: 24 levels; model: 11 levels); random slopes were omitted because models with a maximal random-effects structure did not reliably converge.}
\label{tab:app-gram-pp-pl}
\end{table}

\subsection{ORC, Singular Subject}

\begin{table}[H]
\centering
\small
\begin{tabular}{lrrr}
\toprule
\textbf{Fixed effects} & \textbf{Estimate} & \textbf{Std.\ Error} & \textbf{$t$ value} \\
\midrule
(Intercept)            & 11.8113           & 0.4759               & 24.818 \\
ungram                 & 3.6725            & 0.1384               & 26.531 \\
attr                   & 0.1168            & 0.1384               & 0.844 \\
ungram $\times$ attr   & -1.0312           & 0.1958               & -5.268 \\
\bottomrule
\end{tabular}
\caption{Regression results for object-extracted relative clauses, singular subject.  Observations: 2112.  Random effects: by-item and by-model random intercepts (item: 48 levels; model: 11 levels); random slopes were omitted because models with a maximal random-effects structure did not reliably converge.}
\label{tab:app-gram-orc-sg}
\end{table}

\subsection{ORC, Plural Subject}

\begin{table}[H]
\centering
\small
\begin{tabular}{lrrr}
\toprule
\textbf{Fixed effects} & \textbf{Estimate} & \textbf{Std.\ Error} & \textbf{$t$ value} \\
\midrule
(Intercept)            & 10.9580           & 0.3971               & 27.593 \\
ungram                 & 7.2830            & 0.1140               & 63.869 \\
attr                   & -0.8139           & 0.1140               & -7.137 \\
ungram $\times$ attr   & 0.2673            & 0.1613               & 1.657 \\
\bottomrule
\end{tabular}
\caption{Regression results for object-extracted relative clauses, plural subject.  Observations: 2112.  Random effects: by-item and by-model random intercepts (item: 48 levels; model: 11 levels); random slopes were omitted because models with a maximal random-effects structure did not reliably converge.}
\label{tab:app-gram-orc-pl}
\end{table}

\section{Global-level Regression Analysis: Full Results}
\label{sec:appendix-globalregression}

The following table reports model coefficients for predictors with standard errors. \\

{\noindent \small
\begin{tabular}{lrr}
\toprule \\[-1.8ex]
Predictor                          & Experiment~1           & Experiment~2 \\
\midrule
Intercept                          & 5.512$^{***}$ (0.356)   &  11.811$^{***}$ (0.385) \\
npar                               & $-$0.124 (0.387)       & $-$1.576$^{***}$ (0.399) \\
attr                               & $-$0.114 (0.096)       & 0.117 (0.131) \\
ungram                             & 7.220$^{***}$ (0.096)   & 3.672$^{***}$ (0.131) \\
pl                                 & 0.042 (0.096)          & $-$0.853$^{***}$ (0.131) \\
npar:attr                          & 0.091 (0.194)          & 0.176 (0.264) \\
npar:ungram                        & 1.495$^{***}$ (0.194)   & 0.035 (0.264) \\
npar:pl                            & $-$0.031 (0.194)       & 0.355 (0.264) \\
attr:ungram                        & $-$2.262$^{***}$ (0.136) & $-$1.031$^{***}$ (0.185)\\
attr:pl                            & 0.084 (0.136)          & $-$0.931$^{***}$ (0.185) \\
ungram:pl                          & $-$1.611$^{***}$ (0.136) & 3.611$^{***}$ (0.185)  \\
npar:attr:ungram                   & $-$0.441 (0.275)       & $-$0.111 (0.373) \\
npar:attr:pl                       & $-$0.026 (0.275)       & $-$0.495 (0.373)\\
npar:ungram:pl                     & 0.335 (0.275)          & 0.482 (0.373) \\
attr:ungram:pl                     & 1.917$^{***}$ (0.193)   & 1.299$^{***}$ (0.261) \\
npar:attr:ungram:pl \hspace*{-2em} & 0.262 (0.389)          & 0.717 (0.527) \\
\midrule
Observations & 2,112 & 4,224 \\
\bottomrule
\multicolumn{2}{l}{\textit{Note:} $^{*}$p$<$0.1; $^{**}$p$<$0.05; $^{***}$p$<$0.01} \\
\end{tabular}}

\vspace{1ex}
{\noindent \small Random effects (both experiments): Due to convergence issues, the random effects structure included only by-item and by-model random intercepts (model: 11 levels, item: 48 levels).  Models were not nested within their families: the nested structure yielded a singular fit, presumably because the small number of models does not permit separating model-level from family-level variance.}

\end{document}